\documentclass[conference]{IEEEtran}
\IEEEoverridecommandlockouts
\usepackage{cite}
\usepackage{amsmath,amssymb,amsfonts}
\usepackage{algorithm,algorithmic} 
\usepackage{graphicx}
\usepackage{textcomp}
\usepackage{xcolor}
\usepackage{subcaption}
\usepackage{booktabs}
\usepackage{amsmath,amssymb,amsthm,mathtools}
\usepackage{breqn}
\usepackage{bm}
\usepackage[bookmarks=false]{hyperref}
\usepackage[capitalize]{cleveref} 

\newcommand\bbE{\ensuremath{\mathbb{E}}} 

\def\BibTeX{{\rm B\kern-.05em{\sc i\kern-.025em b}\kern-.08em
		T\kern-.1667em\lower.7ex\hbox{E}\kern-.125emX}}

\begin{document}

	\title{Injective State-Image Mapping facilitates Visual Adversarial Imitation Learning}

	\author{Subhajit Chaudhury, Daiki Kimura, Asim Munawar, Ryuki Tachibana, \\
		{\{subhajit, daiki, asim, ryuki\}@jp.ibm.com}\\
		IBM Research AI}
	\maketitle
	\IEEEpubidadjcol
	\begin{abstract}
		
		The growing use of virtual autonomous agents in applications like games and entertainment demands better control policies for natural-looking movements and actions. Unlike the conventional approach of hard-coding motion routines, we propose a deep learning method for obtaining control policies by directly mimicking raw video demonstrations. Previous methods in this domain rely on extracting low-dimensional features from expert videos followed by a separate hand-crafted reward estimation step. We propose an imitation learning framework that reduces the dependence on hand-engineered reward functions by jointly learning the feature extraction and reward estimation steps using Generative Adversarial Networks~(GANs).  Our main contribution in this paper is to show that under injective mapping between low-level joint state (angles and velocities) trajectories and corresponding raw video stream, performing adversarial imitation learning on video demonstrations is equivalent to learning from the state trajectories. Experimental results show that the proposed adversarial learning method from raw videos produces a similar performance to state-of-the-art imitation learning techniques while frequently outperforming existing hand-crafted video imitation methods. Furthermore, we show that our method can learn action policies by imitating video demonstrations on YouTube with similar performance to learned agents from true reward signals. Please see the supplementary video submission at \url{https://ibm.biz/BdzzNA}.
	\end{abstract}
	
	
	\section{Introduction}
	
	Learning to control artificial agents in simulated environments has potential applications in numerous fields like computer graphics, human-computer interaction, etc. For instance, it can be possible to teach a control policy for artificial agents in games from expert video demonstrations of other humans without hard-coding the specific rules to achieve a certain task. Traditionally, such learning-from-demonstration algorithms learn from low-level information like joint angles and velocities, which have to be carefully acquired from the expert agent. A more natural way for imitation learning involves directly learning from visual demonstrations of the expert without access to such low-level information. In this paper, we present an imitation learning algorithm that can mimic raw videos directly and show that under certain conditions learning directly from raw videos is equivalent to learning from low-level information.
	
	Prior methods in this field have focused on learning from expert demonstration~ \cite{schaal1997learning} or imitation learning~\cite{pomerleau1991efficient,Ng:2000:AIR:645529.657801, ho2016generative}. In most conventional methods, a set of expert trajectories consisting of both state and action information is available which is used to mimic the expert behavior by maximum likelihood estimation. Attempts have also been made towards learning from observations in the absence of action information from low-level motion capture data~\cite{merel2017learning, peng2018deepmimic}. Existing methods for learning from videos focus on extracting a high-level feature embedding of the image frames ensuring that pair of co-located frames lie close in the embedding space compared to the ones having higher temporal distance. This is then followed by a reward estimation step based on some hand-crafted distance metric between the feature embeddings of the agent and expert images. Additionally, these methods make a restrictive assumption that the agent's trajectories need to be time-synchronized to the expert for reward estimation.
	
\begin{figure}[tb]
		\centering
		\includegraphics[width=0.5\textwidth]{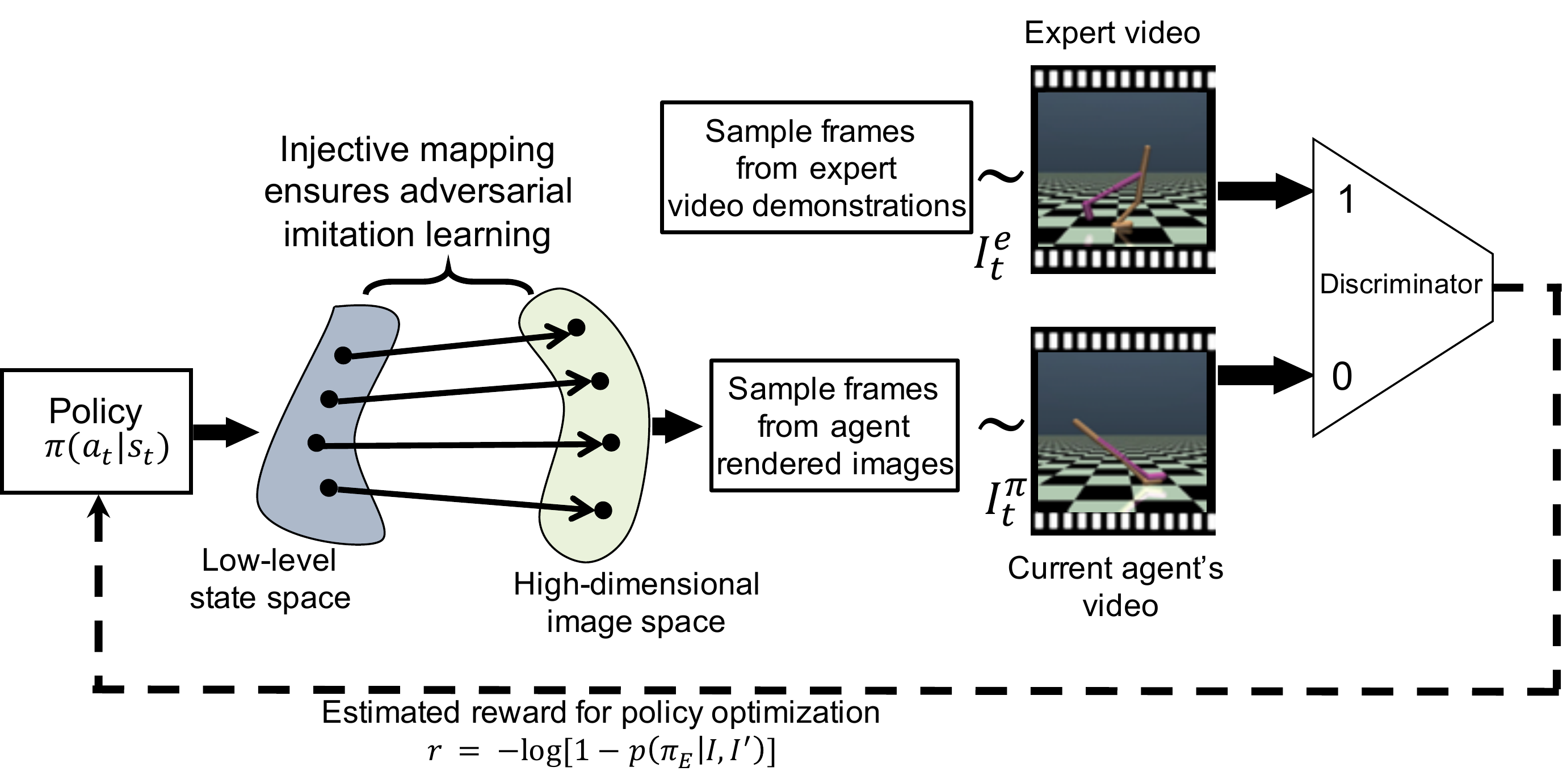}
		\caption{Proposed imitation learning from expert videos}
		\label{fig:overview}
	\end{figure}

	Our adversarial imitation learning method addresses both of the above issues faced in conventional methods. Firstly, we jointly learn the high-level feature embeddings and agent-expert distance metric removing the need for hand-crafted reward shaping. Specifically, we use a generative adversarial network (GAN, \cite{goodfellow2014generative}) framework with the policy network as the generator along with a discriminator, which performs binary classification between the agent and expert trajectories. The reward signal is extracted from the discriminator output indicating the discrepancy between the current performance of the agent with the expert. The overview of our method is shown in Figure \ref{fig:overview}. Secondly, since the binary classifier is trained by randomly sampling the agent and expert trajectories, there is no need for time synchronization as well.
	
	Our main contribution in this paper is to establish a connection between learning from low-level state trajectories and high-dimensional video frames, which enables us to develop a visual adversarial imitation learning algorithm. Specifically, we show that if there exists an injective mapping between the state trajectories and video frames, adversarial learning from raw videos is equivalent to learning from state trajectories. Thus, adversarial methods suitable for learning from low-level state trajectories can be also applied to policy learning from raw videos. This provides the theoretical background supporting our proposed visual imitation algorithm.
	
	Empirical evaluations show that our proposed method can successfully imitate raw video demonstrations producing a performance similar to state-of-the-art imitation learning techniques even though it uses both state and action information in expert trajectories.  Furthermore, we empirically show that the policies learned by our method outperforms other video imitation methods and is robust in the presence of noisy expert demonstrations. Lastly, we demonstrate that the proposed method can successfully learn policies that imitate video demonstrations available on YouTube.

	\section{Notations}
	We use the standard Inverse Reinforcement Learning~(IRL) framework to denote our notations. We consider a finite horizon Markov Decision Process (MDP), defined as $(\mathcal{S}, \mathcal{A}, P, r, \gamma, p_0)$, where $\mathcal{S}$ is the finite state space, $\mathcal{A}$ is the finite action space, $P : \mathcal{S} \times \mathcal{A} \times \mathcal{S} \rightarrow [0, 1]$ is the transition probability, $r : (\mathcal{S} \times \mathcal{A}) \rightarrow \mathbb{R}$ is the reward function that is used to learn the policy during reinforcement learning, $p_0 : \mathcal{S} \rightarrow [0, 1]$ is the distribution of the initial state $s_0$, and $\gamma \in (0, 1)$ is the discount factor. Throughout the paper, we denote an instance, $s \in \mathcal{S}$ as the agent's states containing low-level information like joint angles and velocities. 
	For every state instance, there is a corresponding visual observation $I \in \mathcal{I}$ depicting the agent's state in raw pixels.
	We consider $\pi : \mathcal{S} \times \mathcal{A} \rightarrow [0, 1]$ to be a stochastic policy that estimates a conditional probability of actions given the state at any given time-step. The expected discounted reward (value function) is denoted as $\bbE_{\pi}[r(s,a)] \triangleq \bbE\left[\sum_{t=0}^\infty \gamma^t r(s_t,a_t)\right]$, where $s_0 \sim p_0$, $a_t \sim \pi(\cdot|s_t)$, and $s_{t+1}\sim P(\cdot|s_t,a_t)$ for $t\geq 0$. 
	We assume that some finite number of trajectories $\tau_E$ sampled from an expert policy $\pi_E$ are available and that we cannot interact with the expert policy during imitation learning.

	\section{Background: Adversarial imitation learning}
	\subsection{Generative Adversarial Networks}
	Given some samples from a data distribution $p_{data}$, GANs learn a generator $G(z)$, where $z \sim p_z$ is sampled from the noise distribution, by optimizing the following cost function,
	\begin{equation}
	\begin{aligned}
	\min_{G} \max_ {D \in (0,1)} & \mathbb{E}_{x \sim p_{data}} \log (D(x)) + \\ 
	& \mathbb{E}_{z \sim p_z} \log(1-D(G(z)))
	\end{aligned}
	\end{equation}
	\noindent The above cost function defines a two player game, where the discriminator tries to classify the data generated by the generator as label 0 and samples from the true data distribution as 1. The discriminator acts as a critic that measures how much the samples generated by the generator matches the true data distribution. The generator is trained by assigning label 1 to the generated samples from G, with fixed discriminator. Thus the generator tries to fool the discriminator into believing that the generated samples are from the true data distribution.
	
	GANs provide the benefit that it automatically learns an appropriate loss between the data and generated distributions. The gradients from the above two-step training methods of the binary classifier are sufficient to produce a good generative model even for high dimensional distributions. Deep Convolutional GANs ( \cite {radford2015unsupervised, zhu2017unpaired}) have shown impressive results on learning the distribution of natural images. Therefore in our paper, we use them for imitation learning from video demonstrations.

	\subsection{ Generative Adversarial Imitation Learning}
	Unlike the task of learning a generative model for images, where the different samples can be assumed as i.i.d., the distribution of state-action tuple at each time-step, in MDPs, are conditionally dependent on past values. As such, the visitation frequency of state-action pairs for a given policy is defined as the occupancy measure, $\rho_\pi(s,a) = \pi(a|s) \sum_{t=0}^\infty \gamma^t P(s_t=s|\pi)$. It was shown by~\cite{syed2008apprenticeship} that the imitation learning problem (matching expected long term reward, $\bbE_{\pi}[r(s, a)]$) can be reduced to the occupancy measure matching problem between the agent's policy and the expert. Employing maximum entropy principle to the above occupancy matching problem, we get a general formulation for imitation learning as,
	\begin{equation}
	\label{eq:occMatch}
	\min_{\pi\in\Pi} \;\;\; \bm{d}(\rho_{\pi}(s,a), \rho_{\pi_E}(s,a))-\lambda H(\pi) 
	\end{equation}
	
	\noindent, where $\rho_{\pi_E}$ is the occupancy measure for the expert and $\bm{d(., .),}$ is the distance function between expert and agent's occupancy measure. $H(\pi)$ is the entropy of the policy and $\lambda$ is the weighting factor. Generative adversarial imitation learning~\cite{ho2016generative} proposed to minimize the Jensen-Shannon divergence between the agent's and expert's occupancy measure and showed that this can be achieved in an adversarial setting by finding the saddle point $(\pi, D)$ of the following cost function. 
	
	\begin{multline}
	\label{eq:GAIL}
	\min_{\pi \in \Pi} \max_ {D \in (0,1)} \mathbb{E}_{\pi_E}[\log D(s,a)] + \\ \mathbb{E}_{\pi} [\log(1-D(s,a))] -\lambda H(\pi)
	\end{multline}
	
	\noindent The discriminator $D(s,a) = p(E|s,a)$, represents the likelihood that state-action tuple is generated from the expert rather than by the agent. $\{E, A\}$ are two classes of the binary classifier representing expert and agent. As this two-player game progresses, the discriminator learns to better classify the expert trajectories from the agent and the policy learns to generate trajectories similar to the expert.

	\begin{figure}[tb]
		\centering
		\includegraphics[width=0.4\textwidth]{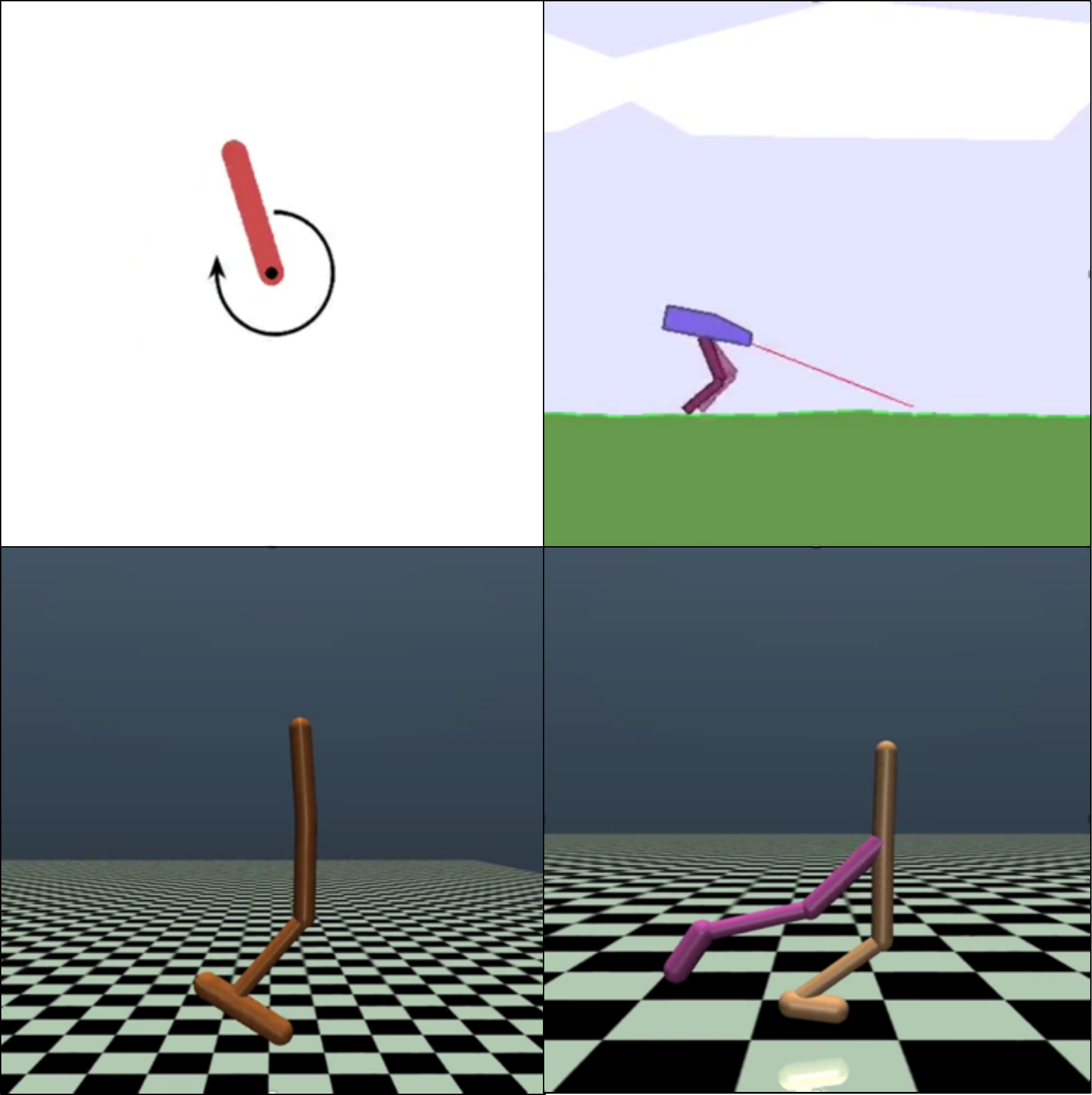}
		\caption{Environments used in our experiments. From top left to bottom right : Pendulum, Bipedal Walker, Hopper, Walker2d}
		\label{fig:envs}
	\end{figure}
	
	\subsection{Learning in the absence of expert actions}
	\label{sec:gailobs}
	Similar to GAIL's objective for imitation learning from expert state-action information, it has been shown~\cite{merel2017learning, henderson2017optiongan}) that it is possible to imitate an expert policy from state demonstrations only, even without action information. Similar to the case of GAIL, we define a reward function $r(s,s')$ which depends only on state transitions, $(s, s')$. We define the $2$-step state transition occupancy measure as, $\rho_{\pi}(s ,s') = \sum_{a} p(s'|a,s) \pi(a|s) \sum_{t=0}^\infty \gamma^t p(s=s_t|\pi) $.
	
	Following similar arguments outlined in~\cite{ho2016generative, torabi2018generative} and replacing state-action pair with state-transition $(s,s')$, we obtain similar conclusions that minimizing the Jensen-Shannon divergence between occupancy measures of the state transitions for the expert and agent, leads to an optimal policy that imitates the expert. We arrive at cost function similar to Eq \ref{eq:GAIL} with state-action pair$(s,a)$ replaced by state transitions $(s,s')$.
	
	The discriminator $D(s,s') = p(E|s,s')$, tries to discriminate between the state-transition samples of the expert and agent. This serves as a reward signal for training the policy $\pi$ using policy gradient method similar to GAIL.
	
	\section{Proposed method}
	
	We assume that only a finite set of expert demonstrations, $\tau^i_E = \{I^i_0, I^i_1, ..., I^i_T\}$, are available consisting of videos depicting the expert's policy behavior in raw pixels. Similar to natural imitation learning setting for humans, the goal is to learn a low-level control policy $\pi(a|s)$ that performs actions based on state representations, by maximizing the similarity between the agent's video output with that of the expert.
	
	\subsection{Relation between learning from low-level states and raw video}
	\label{sect:proof}
	
	We first establish a relationship between learning from states and the corresponding rendered images, that forms the backbone of our video imitation algorithm. Let us assume that $g: \mathcal{S} \rightarrow \mathcal{I}$ is the render mapping, that maps low-level states $s$ to the high-level image observations $I$. This can depend on various factors in the environment like camera angle, agent morphology, other objects, etc. We make the following proposition,
	
	\textbf{\textit{Proposition}}: Matching the data distribution of the images generated by the agent with the expert demonstrations by optimizing the cost function in Equation \ref{eq:VIGAN}, is equivalent to matching the low-level state occupancy measure, given that the mapping $g(.)$ is injective. 
	
	\begin{multline}
	\label{eq:VIGAN}
	\min_{\pi \in \Pi} \max_ {D \in (0,1)} \mathbb{E}_{\pi_E}[\log D(I,I')] +  \mathbb{E}_{\pi} [\log(1-D(I,I'))]
	\end{multline}
	
	\noindent where the discriminator $D(I,I') = p(E|I,I')$ gives the likelihood that the image transition $\{I,I'\}$, is generated by the expert policy $\pi_E$ rather than the agent policy $\pi$.
	
	\textbf{Proof}: We start by restricting the image space to a manifold $\mathcal{I}$ constrained by the state space, such that, $\mathcal{I} = \{g(s),\;\; \forall s \in \mathcal{S}\}$. Since, we assume that $\mathcal{S}$ is a finite state space (similar to \cite{ho2016generative}) and $g$ is an injective mapping, then it follows that $g: \mathcal{S} \rightarrow \mathcal{I}$ is bijective. 
	
	Using Bayes rule, we can decompose $D(I,I')$ as,
	\begin{equation}
	\label{eq:bayes}
	\begin{aligned}
	D(I,I') &= p(E|I,I')=\frac{p(I,I'|E)p(E)}{p(I,I')} \\
	& = \frac{p(I,I'|E)p(E)}{p(I,I'|A)p(A)+p(I,I'|E)p(E)} \\
	&= \frac{1}{1 + \frac{p(I,I'|A)}{p(I,I'|E)}} = \frac{1}{1 + \beta(I,I')}
	\end{aligned}
	\end{equation}
	
	\noindent where $\beta(I,I')=\frac{p(I,I'|A)}{p(I,I'|E)}$ is the likelihood ratio of the image transitions given current agent's policy to that of the expert policy.  We assume that the equal samples from agent and expert are used during the training, $p(E) =  p(A)$. 
	
	Since there exists a bijective mapping between the high-level image space($\bm{I}$) and the low-level state space ($\bm{s}$), we can write 
	\begin{equation}
	\label{eq:jacobian}
	\begin{aligned}
	\beta(I,I') = \frac{p(I,I'|A)}{p(I,I'|E)} = \frac{p(s,s'|A) |\det J|^{-1}}{p(s,s'|E)|\det J|^{-1}} = \frac{p(s,s'|A)}{p(s,s'|E)}
	\end{aligned}
	\end{equation}
	
	where $J=\frac{\bm{dI}}{\bm{ds}}$ is the Jacobian of bijective function $\bm{g}$. 
	\footnote{Since, we constrain the image space to a manifold governed by the state space, the rank of image manifold will be same as the dimensionality of state space. Thus, we can find maximally independent set of rows for J to construct a full rank Jacobian matrix.}
	\footnote{To be precise, for two step image transitions, $I,I'$, injectivity of the concatenated mapping, $f([s;s'])=[g(s);g(s')]=[I;I']$, is sufficient. However, injectivity of $g$ is a stronger assumption and implies injectivity of $f$.} 
	From Equation \ref{eq:jacobian}, it follows that $D(I,I')=D(s,s')$.
	Therefore, learning a discriminator $D(I,I')$ on the image space, is equivalent to learning a binary classifier on the state space that distinguishes between the expert and agent trajectories. Thus, using an estimated reward function based on the image discriminator for policy optimization would lead to occupancy measure matching in the low-level state space which in turn will learn a policy that imitates the expert, as shown in the previous section. This concludes the proof.\qed 
	
	\subsection{Interpretation of the injective mapping assumption}
	The above analysis states that an injective mapping between low-level states and high dimensional image enables adversarial imitation learning from raw videos. This assumption is reasonable for most environments where the agent resides in a two-dimensional world. For agents residing in a $3$D world, there can be some degenerate cases where this assumption is not valid. Since images capture a $2$D projection of the $3$D world, one such case can include moving along the axis of the camera with increasing size (such that the projection on the camera does not change). Another case can be when the agent is occluded by some object in the $3$D world such that changing the agent's state does not register any change in the video frames. At those degenerate points, the Jacobian $J$ will be singular and Equation \ref{eq:jacobian} will not be valid. We perform an empirical evaluation on both 2D and 3D environments and show that if the expert trajectory does not contain such rare degenerate cases, our proposed method can recover a favorable policy.

	\begin{figure}
		
		\centering
		\begin{subfigure}[b]{0.24\textwidth}
			\includegraphics[width=0.98\linewidth]{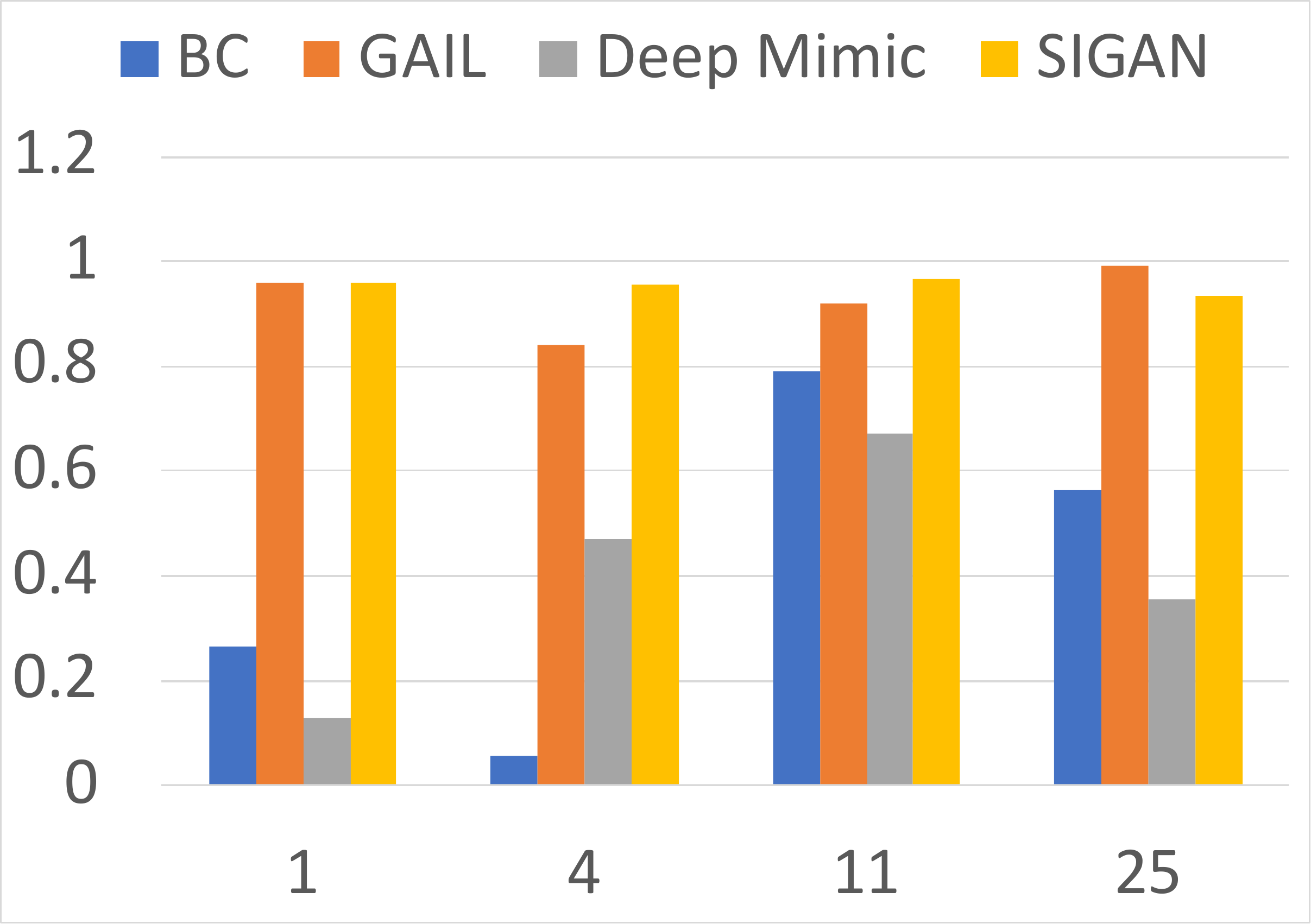}
			\caption{Hopper}
			\label{fig:SIGAN-Hopper}
		\end{subfigure}%
		\begin{subfigure}[b]{0.24\textwidth}
			\includegraphics[width=0.98\linewidth]{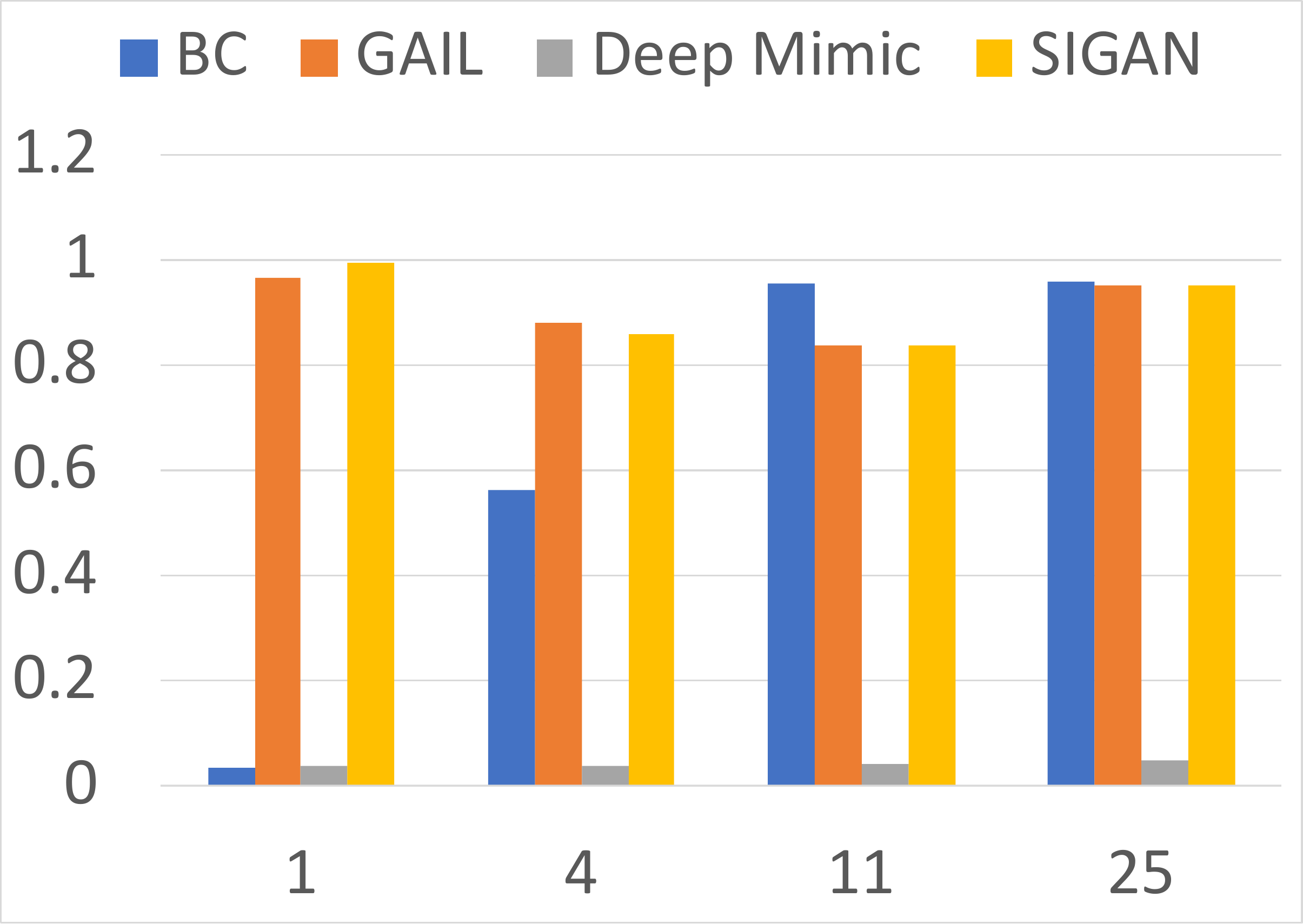}
			\caption{Walker2d}
			\label{fig:SIGAN-Walker}
		\end{subfigure}%
		\caption{Imitation learning from low-level state stajectories. BC and GAIL use expert state-action tuples, while DeepMimic and SIGAN use only state information. SIGAN performs similarly as GAIL while being superior in performance to BC and DeepMimic. Higher reward suggests better policy.}\label{fig:SIGAN}
	\end{figure}

	\subsection{Algorithm for adversarial imitation learning from video}
	
	We outline the practical algorithm for Video Imitation Generative Adversarial Network (VIGAN) as algorithm \ref{alg:VIGAN}.
	
	\begin{algorithm}[h]
		\caption{$VIGAN(\tau_E)$}
		\label{alg:VIGAN}
		\begin{algorithmic}[1]
			\REQUIRE $\tau_E$: Expert video demonstrations without action
			\STATE - Randomly initialize the parameters $\bm{\theta}$ for policy $\pi$ and $\bm{\phi}$ for the discriminator $D$
			\FOR{$k$ from $0$ to $\infty$, until convergence}
			\STATE Execute $\pi_{\theta}$ and store the state transition $\{s, a, s'\} \rightarrow \tau_s$ for $T$ time-steps.
			\STATE Render corresponding raw images $I=g(s)$ and store generated video ($I_1, I_2,...,I_T$) in image buffer, $\tau_I$
			\STATE Perform a gradient step for discriminator parameters from $\bm{\phi}_{k}$ to $\bm{\phi}_{k+1}$ using loss as, \\
			$-\hat{\mathbb{E}}_{\tau_E}[\log D_{\phi}(I,I')] - \hat{\mathbb{E}}_{\tau_I} [\log(1-D_{\phi}(I,I'))$
			\STATE Estimate reward from the discriminator and use it for policy otimization as,\\ $r_t = -\log(1-D_{\phi_{k+1}}(I_t,I_{t+1}))$
			\ENDFOR
			\STATE Return $\pi_{\theta}$
		\end{algorithmic}
	\end{algorithm}
	
	Our algorithm directly uses consecutive video frames instead of on low-level states for training the discriminator according to the proposition made in Section \ref{sect:proof}. We use TRPO~\cite{schulman2015trust} for policy optimization and variance reduction in policy gradients is performed following \cite{schulman2015high}.
	
	Our GAN-based reward estimation automatically learns both feature representation from images and the distance between such embeddings in a joint fashion from the data-distribution of expert and the agent. This reduces the need for hand-crafted reward shaping. Secondly, since we train the discriminator by comparing random samples drawn from both agent and expert distributions, there is no need for them to be time-synchronized. Thus, our method addresses both issues of hand-crafted reward estimation and time-synchronized agent-expert matching faced by previous works.

	\section{Experiments}
	We perform four quantitative evaluations to validate our claim: (1) Perform imitation learning from low-level state only without action from expert, similar to~\cite{ho2016generative, torabi2018generative}, which we refer to as State Imitation Generative Adversarial Networks (SIGAN), (2) Evaluate the proposed VIGAN algorithm from raw videos demonstrations on standard environments, (3) Demonstrate the robustness of our proposed method to noise, and (4) Learn from YouTube video demonstrations. Qualitative comparisons are shown in the video submission.
	
	A neural network policy, consisting of 2 fully connected layers of 64 ReLU activated units each, was used for all experiments. For the discriminator, we used a convolution neural network(CNN) with similar architecture as in \cite{radford2015unsupervised} for video imitation. 
	We evaluate our algorithm on four physics-based environments: two classical control tasks of CartPole and Pendulum, along with 3 continuous control tasks of Hopper, Walker2d, and BipedalWaker from OpenAI's gym environment~\cite{brockman2016openai} as shown in Figure \ref{fig:envs}.

	\subsection{Imitation from Low-Level States}
	
	In this experiment, we perform experiments to show that adversarial learning in the absence of actions~\cite{ho2016generative, torabi2018generative}, as mentioned in Section \ref{sec:gailobs}, performs similarly to state-of-the-art imitation learning methods. We compared the performance of SIGAN for 3 different values of $n$ with behavior cloning, GAIL and DeepMimic (which uses heuristic rewards) as shown in Figure \ref{fig:SIGAN}. Results show that adversarial state imitation can recover a policy that performs at-par with GAIL and out-performs other methods.

	\subsection{Imitation from Raw Videos}
	Having demonstrated that learning from observations performs similar to GAIL~\cite{ho2016generative}, we perform experiments to show that the proposed equivalence between learning from state trajectories and video frames is valid. Our experiments demonstrate that video imitation can also perform similarly with state-of-the-art imitation learning methods. We use expert trajectories of the video demonstrations consisting of 800 image frames for Hopper and Walker2d environments. CartPole and Pendulum used 200 frames per trajectory. Both the rendered videos from expert and the agent were resized to $64 \times 64 \times 3$ color images for training the discriminator. 
	
	\begin{table}[t]
		\small
		\begin{center}
			\begin{tabular}{cccccc}
				\toprule
				Env & \#Traj &  GAIL & DM+Pix & DM+TCN & Proposed \\\midrule
				CartPole & 1 & $\textit{200.0} $  & $191.6 $ & $200.0$ & $200.0$\\ 
				& 5 & $\textit{200.0}$  & $200.0$ & $200.0$ & $200.0$\\ 
				& 10 & $\textit{200.0}$  & $200.0$ & $200.0$ & $200.0$\\\midrule
				
				Pendulum & 1 & $\textit{-242.0}$  & $-1203.4$ & $-732.3$ & $\bm{-194.4}$\\ 
				& 5 & $\textit{-278.6}$  & $-1247.9$ & $\bm{-205.6}$ & $-287.5$\\ 
				& 10 & $\textit{-313.0}$  & $-1298.3$ & $-209.2$ & $\bm{-177.4}$\\\midrule
				
				Hopper & 1 & $\textit{3607.1}$  & $1012.1$ & $619.3$ & $\bm{3053.6}$\\ 
				& 4 & $\textit{3159.6} $  & $1008.3$ & $610.6$ & $\bm{2513.3}$\\ 
				& 11 & $\textit{3466.6}$  & $979.8$ & $624.5$ & $\bm{2490.3}$\\ 
				& 25 & $\textit{3733.5}$  & $990.0$ & $605.5$ & $\bm{2812.2}$\\\midrule
				
				Walker2d & 1  & $\textit{5673.6}$ & $537.3$ & $747.8$ & $\bm{2505.9}$\\ 
				& 4 & $\textit{5160.5}$ & $729.9$   & $3659.5$ & $\bm{4211.2}$\\ 
				& 11 & $\textit{4920.7}$ & $846.3$   & $629.4$ & $\bm{4340.4}$\\ 
				& 25 & $\textit{5596.6}$ & $495.5$  & $3356.4$ & $\bm{5606.4}$\\        
				\hline
			\end{tabular}
			\caption{Final policy performance learned by various video imitation methods during inference. Our proposed method out-performs existing video imitation methods while producing a similar performance to GAIL. It is to be noted here that GAIL was trained from low-level state-action tuples while other methods used only video demonstrations without actions. DM is ``DeepMimic" and ``TCN" is Time Contrastive Networks.}
			\label{table:results}
		\end{center}
	\end{table}
	
	We compare our proposed method to the following prior works in video imitation methods in addition to GAIL\cite{ho2016generative}. 
	
	\textbf{DeepMimic~\cite{peng2018deepmimic} + PixelLoss} : In this method, the reward was simply computed as Euclidean distance between normalized images (in the range [-1, 1]) rendered from the agent and expert. We found that taking exponential of the distance, provides a more stable performance. Thus the reward at time $t$ is computed as, $r_t = \exp(-2(||I^e_t - I^{\pi}_t||_2^2))$, where $I^e_t$ and $I^{\pi}_t$ are normalized images sampled from the expert demonstrations and agent policy, respectively.
	
	\textbf{DeepMimic~\cite{peng2018deepmimic} + Single View TCN~\cite{sermanet2017time}} : We used Single View TCN for self-supervised representation learning using implementation of triplet loss provided by the author. The triplet loss encourages embeddings for co-located images to lie close to each other while separating embeddings for images that are not semantically related. The reward at time $t$ was computed as $r_t = \exp(-2||x^e_t - x^{\pi}_t||_2^2)$, where $x^e_t$ and $x^{\pi}_t$ are the agent's and expert's render images.
	
	Quantitative evaluation for all the methods is given in Table \ref{table:results}. For DeepMimic, learning from raw pixel information did not provide good performance for complex environments because it does not capture the high-level semantic information about the agent's state. Single View TCN + DeepMimic performed well in some cases (Walker2d, Pendulum) but did not consistently produce a good performance for all cases. We believe that using reward shaping with careful parameter tuning might lead to improved performance for TCN. Our method consistently performed better than the other video imitation methods and was comparable to GAIL's performance which was trained on both state and action trajectories. Furthermore, our reward estimator does not need much parameter tuning (only the number of discriminator steps per iteration was varied) across environments, because VIGAN automatically finds the separation between agent's current video trajectories to that of the expert, ensuring robust estimation of the reward signal from raw pixels.
	
	\subsection{Imitation from Noisy Videos}
	
	In this experiment, we added noise to the input video demonstrations to evaluate the robustness of the proposed algorithm to small viewpoint changes. The noise was added in the form of a shaking camera, which was simulated by randomly cropping each video frame by 0 to 5\% from all four sides. Such noisy video demonstrations might break the injective nature of render mapping because the same low-level state might be mapped to different image observations.
	
	We found that imitation from noisy demonstrations performed similarly to the non-noisy case and GAIL as shown in Figures \ref{fig:noisyhopper} and \ref{fig:noisywalker}. For the Walker2d environment, we found that learning the discriminator with two image transitions did not produce a good performance. Therefore we used 3 consecutive image transitions, $D(I_i,I_{i+1},I_{i+2})$, for training the discriminator which produced better results in comparison. 
	
	\subsection{Imitating YouTube Videos}
	Finally, we used our proposed method to imitate video demonstrations available on YouTube which is the closest to natural imitation learning in humans. We chose the BipedalWalker environment from OpenAI gym because we found two videos {\footnote{Link: {https://www.youtube.com/watch?v=uwz8JzrEwWY}} \footnote{Link: {https://www.youtube.com/watch?v=nWd2cN5oriM}}} for this environment with different walking styles trained by others. 
	Since the raw expert video demonstrations from YouTube contained additional artifacts like text and window borders, we cropped the videos to keep the area around the agent's location and resized each frame to $64 \times 64 \times 3$ images for both expert and learning agent.

	\begin{figure*}[tb]
		\centering
		\captionsetup[subfigure]{justification=centering}
		\begin{subfigure}[b]{0.33\textwidth}
			\includegraphics[width=\linewidth]{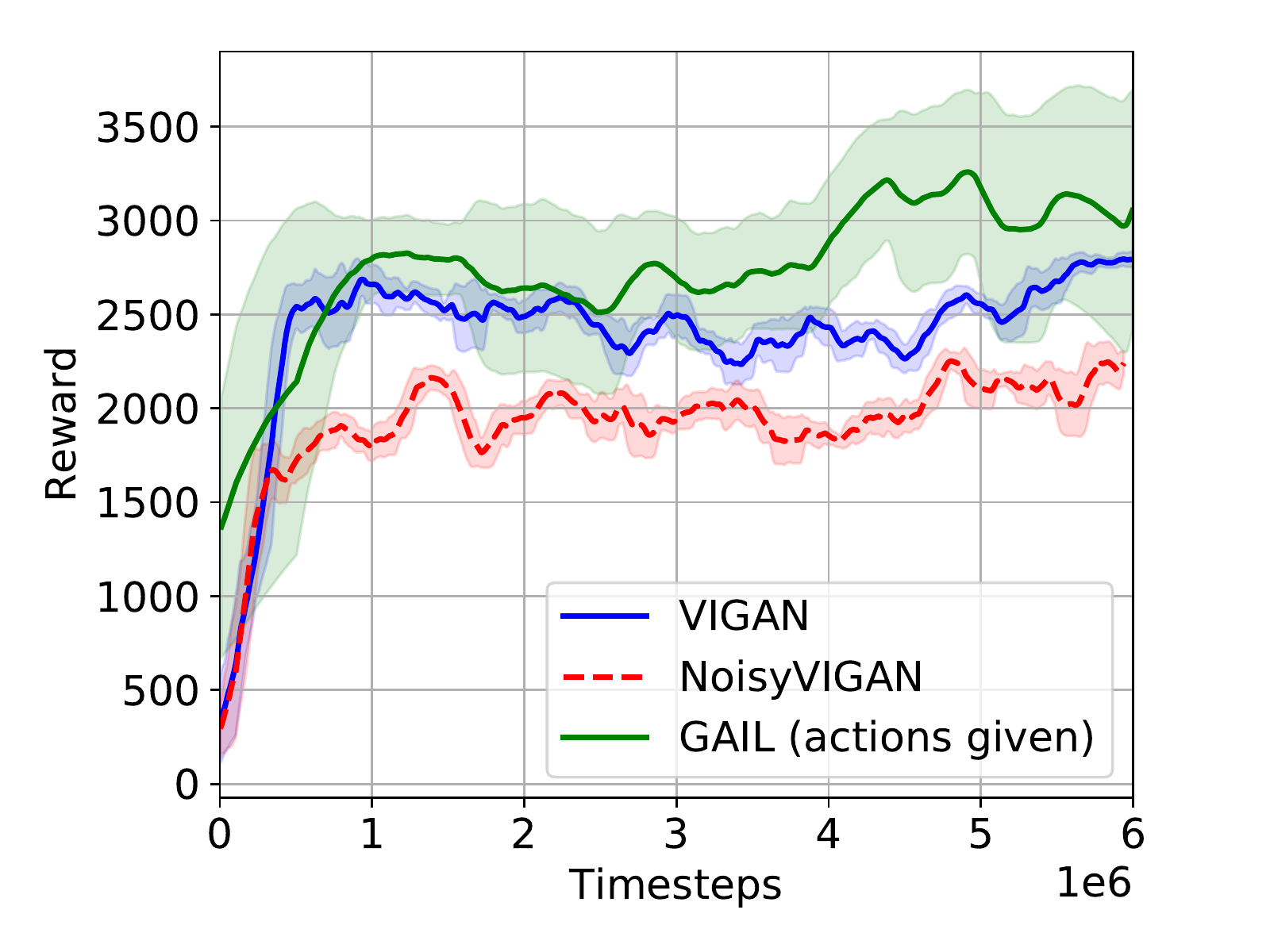}
			\caption{Noisy demonstrations (Hopper)}
			\label{fig:noisyhopper}
		\end{subfigure}%
		\begin{subfigure}[b]{0.33\textwidth}
			\includegraphics[width=\linewidth]{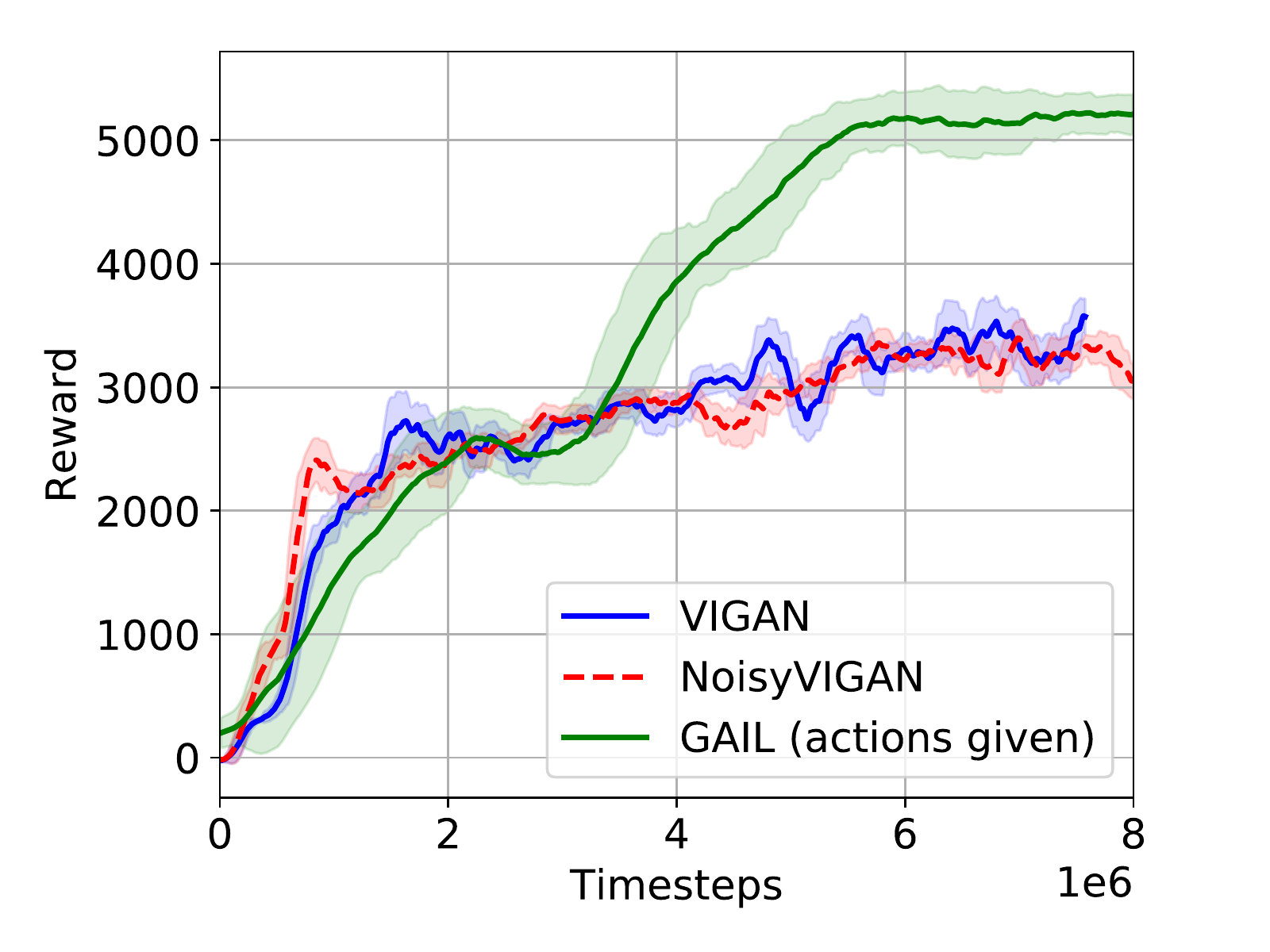}
			\caption{Noisy demonstrations (Walker2d)}
			\label{fig:noisywalker}
		\end{subfigure}%
		\begin{subfigure}[b]{0.33\textwidth}
			\includegraphics[width=\linewidth]{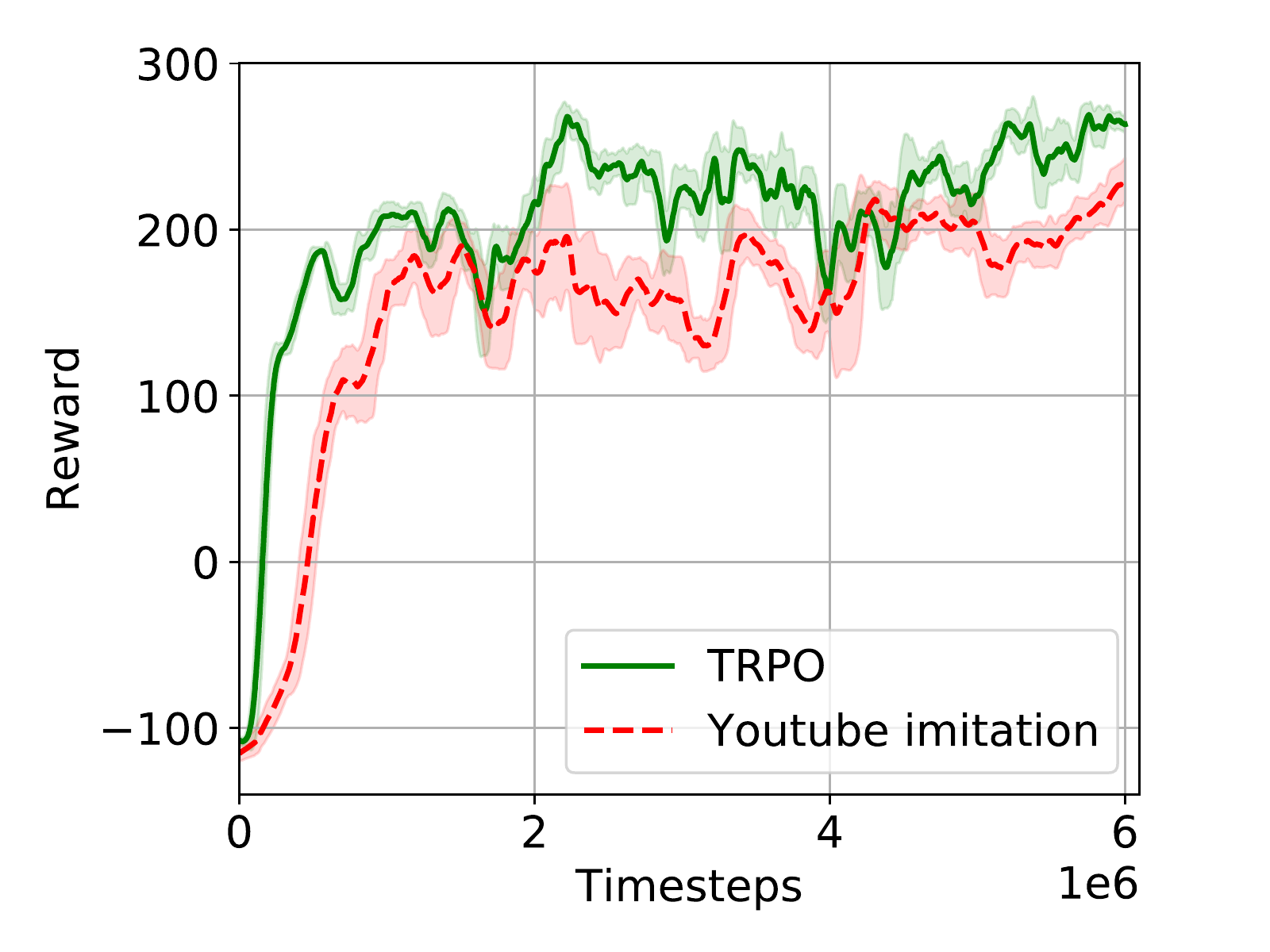}
			\caption{Learning from YouTube video}
			\label{fig:youtube}
		\end{subfigure}%
		
		\caption{(a) VIGAN's policy performance with noisy video demonstrations for Hopper environment is comparable to the non-noisy case showing the robustness of the proposed method. Training was done from 1 expert trajectory (b) Robustness of VIGAN to noise in Walker2d environment ($4$ expert trajectories) (c) Using VIGAN on YouTube videos produce similar results to an agent trained with TRPO from the true reward. Higher reward suggests better policy.}\label{fig:animals}
	\end{figure*}
	
	We compare the learning curve of our algorithm (for video 2) with an expert agent trained via TRPO using the true reward provided by OpenAI gym~\cite{brockman2016openai}, as shown in Figure \ref{fig:youtube}. Quantitative evaluations show that our method successfully learned a policy, by imitating just a short duration of YouTube videos, that performs at par with the expert policy learned using the true reward signal. It is to be noted that since the expert demonstrations, in this case, were directly taken from YouTube, we had no knowledge of the framerate at which the video was recorded.  For previous methods, that use time-synchronized reward estimation from videos, additional hyperparameters might be required to match the framerate of the expert demonstrations with the agent. However, our method does not require any such time alignment step.
	
	\section{Conclusion}
	We proposed an imitation learning method for recovering control policies from a limited number of raw video demonstrations using generative adversarial video matching for reward estimation. We showed that if there exists an injective mapping between the low-level states and image frames, adversarial imitation from videos and agent-expert state trajectories matching, are equivalent problems. Our proposed method consistently out-performed other video imitation methods and recovered a good policy even in the presence of noise. We further demonstrate that our video imitation method can learn policies imitating youtube videos trained by others. In the future, we would like to extend our method imitate complex video demonstrations with changing background contexts, for example, Montezuma's Revenge or Torcs driving simulator, where hierarchical adversarial reward estimation based on semantic video clustering, would be required.
	
	\section*{Acknowledgment}
	
	We are thankful to Akshay L Aradhya and Anton Pechenko for allowing us to use their YouTube videos for this research project. We would like to thank Jayakorn Vongkulbhisal and Hiroshi Kajino for helpful technical discussions.
	
	\bibliographystyle{IEEEtran}
	\bibliography{vigan}

\begin{thebibliography}{10}
\providecommand{\url}[1]{#1}
\csname url@samestyle\endcsname
\providecommand{\newblock}{\relax}
\providecommand{\bibinfo}[2]{#2}
\providecommand{\BIBentrySTDinterwordspacing}{\spaceskip=0pt\relax}
\providecommand{\BIBentryALTinterwordstretchfactor}{4}
\providecommand{\BIBentryALTinterwordspacing}{\spaceskip=\fontdimen2\font plus
\BIBentryALTinterwordstretchfactor\fontdimen3\font minus
  \fontdimen4\font\relax}
\providecommand{\BIBforeignlanguage}[2]{{%
\expandafter\ifx\csname l@#1\endcsname\relax
\typeout{** WARNING: IEEEtran.bst: No hyphenation pattern has been}%
\typeout{** loaded for the language `#1'. Using the pattern for}%
\typeout{** the default language instead.}%
\else
\language=\csname l@#1\endcsname
\fi
#2}}
\providecommand{\BIBdecl}{\relax}
\BIBdecl

\bibitem{schaal1997learning}
S.~Schaal, ``Learning from demonstration,'' in \emph{Advances in neural
  information processing systems}, 1997, pp. 1040--1046.

\bibitem{pomerleau1991efficient}
D.~A. Pomerleau, ``Efficient training of artificial neural networks for
  autonomous navigation,'' \emph{Neural Computation}, vol.~3, no.~1, pp.
  88--97, 1991.

\bibitem{Ng:2000:AIR:645529.657801}
\BIBentryALTinterwordspacing
A.~Y. Ng and S.~J. Russell, ``Algorithms for inverse reinforcement learning,''
  in \emph{Proceedings of the Seventeenth International Conference on Machine
  Learning}, ser. ICML '00.\hskip 1em plus 0.5em minus 0.4em\relax San
  Francisco, CA, USA: Morgan Kaufmann Publishers Inc., 2000, pp. 663--670.
  [Online]. Available: \url{http://dl.acm.org/citation.cfm?id=645529.657801}
\BIBentrySTDinterwordspacing

\bibitem{ho2016generative}
J.~Ho and S.~Ermon, ``Generative adversarial imitation learning,'' in
  \emph{Advances in Neural Information Processing Systems}, 2016, pp.
  4565--4573.

\bibitem{merel2017learning}
J.~Merel, Y.~Tassa, S.~Srinivasan, J.~Lemmon, Z.~Wang, G.~Wayne, and N.~Heess,
  ``Learning human behaviors from motion capture by adversarial imitation,''
  \emph{arXiv preprint arXiv:1707.02201}, 2017.

\bibitem{peng2018deepmimic}
\BIBentryALTinterwordspacing
X.~B. Peng, P.~Abbeel, S.~Levine, and M.~van~de Panne, ``Deepmimic:
  Example-guided deep reinforcement learning of physics-based character
  skills,'' \emph{ACM Trans. Graph.}, vol.~37, no.~4, pp. 143:1--143:14, Jul.
  2018. [Online]. Available: \url{http://doi.acm.org/10.1145/3197517.3201311}
\BIBentrySTDinterwordspacing

\bibitem{goodfellow2014generative}
I.~Goodfellow, J.~Pouget-Abadie, M.~Mirza, B.~Xu, D.~Warde-Farley, S.~Ozair,
  A.~Courville, and Y.~Bengio, ``Generative adversarial nets,'' in
  \emph{Advances in neural information processing systems}, 2014, pp.
  2672--2680.

\bibitem{radford2015unsupervised}
A.~Radford, L.~Metz, and S.~Chintala, ``Unsupervised representation learning
  with deep convolutional generative adversarial networks,'' \emph{arXiv
  preprint arXiv:1511.06434}, 2015.

\bibitem{zhu2017unpaired}
J.-Y. Zhu, T.~Park, P.~Isola, and A.~A. Efros, ``Unpaired image-to-image
  translation using cycle-consistent adversarial networks,'' in \emph{IEEE
  International Conference on Computer Vision}, 2017.

\bibitem{syed2008apprenticeship}
U.~Syed, M.~Bowling, and R.~E. Schapire, ``Apprenticeship learning using linear
  programming,'' in \emph{Proceedings of the 25th international conference on
  Machine learning}.\hskip 1em plus 0.5em minus 0.4em\relax ACM, 2008, pp.
  1032--1039.

\bibitem{henderson2017optiongan}
P.~Henderson, W.-D. Chang, P.-L. Bacon, D.~Meger, J.~Pineau, and D.~Precup,
  ``Optiongan: Learning joint reward-policy options using generative
  adversarial inverse reinforcement learning,'' \emph{arXiv preprint
  arXiv:1709.06683}, 2017.

\bibitem{torabi2018generative}
F.~Torabi, G.~Warnell, and P.~Stone, ``Generative adversarial imitation from
  observation,'' \emph{arXiv preprint arXiv:1807.06158}, 2018.

\bibitem{schulman2015trust}
J.~Schulman, S.~Levine, P.~Abbeel, M.~Jordan, and P.~Moritz, ``Trust region
  policy optimization,'' in \emph{Proceedings of the 32nd International
  Conference on Machine Learning (ICML-15)}, 2015, pp. 1889--1897.

\bibitem{schulman2015high}
J.~Schulman, P.~Moritz, S.~Levine, M.~Jordan, and P.~Abbeel, ``High-dimensional
  continuous control using generalized advantage estimation,'' \emph{arXiv
  preprint arXiv:1506.02438}, 2015.

\bibitem{brockman2016openai}
G.~Brockman, V.~Cheung, L.~Pettersson, J.~Schneider, J.~Schulman, J.~Tang, and
  W.~Zaremba, ``Openai gym,'' \emph{arXiv preprint arXiv:1606.01540}, 2016.

\bibitem{sermanet2017time}
P.~Sermanet, C.~Lynch, Y.~Chebotar, J.~Hsu, E.~Jang, S.~Schaal, and S.~Levine,
  ``Time-contrastive networks: Self-supervised learning from video,''
  \emph{arXiv preprint arXiv:1704.06888}, 2017.

\end{thebibliography}

\end{document}